\documentclass[11pt]{article}
\usepackage{times}
\usepackage{graphicx}
\usepackage{latexsym}
\usepackage{enumitem}
\usepackage[noend]{algorithmic}
\usepackage{algorithm}
\usepackage{color}
\usepackage{authblk}

\usepackage{latexsym, amssymb, amsmath}

\input epsf

\newtheorem{proposition}{Proposition}

\newenvironment{proof}{{\bf Proof: }}

\newcommand{\hide}[1]{}

\title{Improving the Performance of maxRPC}

\author{Thanasis Balafoutis }
\affil{Department of Information and Communication Systems Engineering, University of the Aegean, Greece. }
\author{Anastasia Paparrizou }
\author{Kostas Stergiou }
\affil{Department of Informatics and Telecommunications Engineering, University of Western Macedonia, Greece. }
\author{Toby Walsh }
\affil{NICTA, University of New South Wales, Australia.}

\date{}

\begin{document}    

\maketitle

\begin{abstract}

Max Restricted Path Consistency (maxRPC) is a local consistency for binary constraints that can achieve considerably stronger pruning than arc consistency. However, existing maxRPC algorithms suffer from overheads and redundancies as they can repeatedly perform many constraint checks without triggering any value deletions. In this paper we propose techniques that can boost the performance of maxRPC algorithms. These include the combined use of two data structures to avoid many redundant constraint checks, and heuristics for the efficient ordering and execution of certain operations. Based on these, we propose two closely related maxRPC algorithms. The first one has optimal O$(end^3)$ time complexity, displays good performance when used stand-alone, but is expensive to apply during search. The second one has O($en^2d^4$) time complexity, but a restricted version with O($end^4$) complexity can be very efficient when used during search. Both algorithms have O$(ed)$ space complexity when used stand-alone. However, the first algorithm has O$(end)$ space complexity when used during search, while the second retains the O$(ed)$ complexity. 
Experimental results demonstrate that the resulting methods constantly outperform previous algorithms for maxRPC, often by large margins, and constitute a more than viable alternative to arc consistency.
\end{abstract}

\section{Introduction}
\label{sec:intro}

%Max Restricted Path Consistency (maxRPC) 
maxRPC is a strong domain filtering consistency for binary constraints introduced in 1997 by Debruyne and Bessiere \cite{db97}. maxRPC achieves a stronger level of local consistency than arc consistency (AC), and in \cite{db01} it was identified, along with singleton AC (SAC), as a promising alternative to AC. Although SAC has received considerable attention since, maxRPC has been comparatively overlooked. The basic idea of maxRPC is to delete any value $a$ of a variable $x$ that has no arc consistency (AC) or path consistency (PC) support in a variable $y$. A value $b$ is an AC support for $a$ if the two values are compatible, and it is also a PC support for $a$ if this pair of values is path consistent. A pair of values $(a,b)$ is path consistent iff for every third variable there exists at least one value, called a PC witness, that is compatible with both $a$ and $b$.

The first algorithm for maxRPC was proposed in \cite{db97}, and two more algorithms have been proposed since then \cite{grandoni03,vion09}. The algorithms of \cite{db97} and \cite{vion09} have been evaluated on random problems only, while the algorithm of \cite{grandoni03} has not been experimentally evaluated at all. Despite achieving considerable pruning, existing maxRRC algorithms suffer from overhead and redundancies as they can repeatedly perform many constraint checks without triggering any value deletions. These constraint checks occur when a maxRPC algorithm searches for an AC support for a value and when, having located one, it checks if it is also a PC support by looking for PC witnesses in other variables. As a result, the use of maxRRC during search often slows down the search process considerably compared to AC, despite the savings in search tree size. 

In this paper we propose techniques to improve the applicability of maxRPC by eliminating some of these redundancies while keeping a low space complexity. We also investigate approximations of maxRPC that only make slightly fewer value deletions in practice, while being significantly faster. We first demonstrate that we can avoid many redundant constraint checks and speed up the search for AC and PC supports through the careful and combined application of two data structures already used by maxRPC and AC algorithms \cite{grandoni03,vion09,BRYZ05,lecoutre07,liki07}. Based on this, we propose a coarse-grained maxRPC algorithm called \texttt{maxRPC3} with optimal O$(end^3)$ time complexity. This algorithm displays good performance when used stand-alone (e.g. for preprocessing), but is expensive to apply during search. We then propose another maxRPC algorithm, called \texttt{maxRPC3$^{rm}$}. This algorithm has O($en^2d^4$) time complexity, but a restricted version with O($end^4$) complexity can be very efficient when used during search through the use of {\em residues}. Both algorithms have O$(ed)$ space complexity when used stand-alone. However, \texttt{maxRPC3} has O$(end)$ space complexity when used during search, while \texttt{maxRPC3$^{rm}$} retains the O$(ed)$ complexity.

Similar algorithmic improvements can be applied to {\em light maxRPC} (lmaxRPC), an approximation of maxRPC \cite{vion09}. This achieves a lesser level of consistency compared to maxRPC but still stronger than AC, and is more cost-effective than maxRPC when used during search. Experiments confirm that lmaxRPC is indeed a considerably better option than maxRPC. 

We also propose a number of heuristics that can be used to efficiently order the searches for PC supports and witnesses.  
Interestingly, some of the proposed heuristics not only reduce the number of constraint checks but also the number of visited nodes. 

We make a detailed experimental evaluation of new and existing algorithms on various problem classes. This is the first wide experimental study of algorithms for maxRPC and its approximations on benchmark non-random problems. Results show that our methods constantly outperform existing algorithms, often by large margins. When applied during search our best method offers up to one order of magnitude reduction in constraint checks, while cpu times are improved up to four times compared to the best existing algorithm. In addition, these speed-ups enable a search algorithm that applies lmaxRPC to compete with or outperform MAC on many problems.

\section{Background and Related Work}
\label{sec:background}

A {\em Constraint Satisfaction Problem} (CSP) is defined as a tuple $(X,D,C)$ where: $X=\{x_1,\ldots,x_n\}$ is a set of $n$ variables, $D=\{D(x_1),\ldots,D(x_n)\}$ is a set of domains, one for each variable, with maximum cardinality $d$, and $C=\{c_1,\ldots,c_e\}$ is a set of $e$ constraints. Each constraint $c$ is a pair $(var(c),rel(c))$, where $var(c)=\{x_{1},\ldots,x_{m}\}$ is an ordered subset of $X$, and $rel(c)$ is a subset of the \textit{Cartesian} product
$D(x_{1})\times \ldots\times D(x_{m})$ that specifies the allowed combinations of values for the variables in $var(c)$. In the following, a binary constraint $c$ with $var(c)=\{x_i,x_j\}$ will be denoted by $c_{ij}$, and $D(x_i)$ will denote the current domain of variable $x_i$.
Each tuple $\tau\in rel(c)$ is an ordered list of values $(a_{1},\ldots,a_{m})$ such that $a_j\in D(x_j)$,$j=1,\ldots,m$.
A tuple $\tau\in rel(c_i)$ is {\em valid} iff none of the values in the tuple has been removed from the domain of the corresponding variable. 

The process which verifies whether a given tuple is allowed by a constraint $c$ is called a {\em constraint check}.
%A constraint $c$ can be either defined {\em extensionally} by explicitly giving $rel(c)$, or (usually) {\em intensionally} by implicitly specifying $rel(c)$ through a predicate or arithmetic function. 
A binary CSP is a CSP where each constraint involves at most two variables. % and is typically represented by a constraint graph where nodes correspond to variables and edges correspond to constraints. 
We assume that binary constraint checks are performed in constant time.
%We denote by $t$ the number of triangles (i.e. 3-cliques) of variables in the constraint graph of a problem. 
%In a binary CSP, a directed constraint $c$, with $var(c)=\{x_i,x_j\}$, is {\em arc consistent} (AC) iff for every value $a_i\in D(x_i)$ there exists a value $a_j\in D(x_j)$ s.t. the 2-tuple $<$$(x_i,a_i),(x_j,a_j)$$>$ satisfies $c$. In this case $(x_j,a_j)$ is called an {\em AC-support} of $(x_i,a_i)$ on $c$. A problem is AC iff there is no empty domain in $D$ and all the constraints in $C$ are AC. 
In a binary CSP, a value $a_i\in D(x_i)$ is {\em arc consistent} (AC) iff for every constraint $c_{ij}$ there exists a value $a_j\in D(x_j)$ s.t. the pair of values $(a_i,a_j)$ satisfies $c_{ij}$. In this case $a_j$ is called an {\em AC-support} of $a_i$. A variable is AC iff all its values are AC. A problem is AC iff there is no empty domain in $D$ and all the variables in $X$ are AC. 

%The {\em revision} of a constraint $c_{ij}$ using a local consistency $A$ is the process of checking whether the values of $x_i$ verify the property of $A$. For example, the revision of $c_{ij}$  using AC verifies if all values in $D(x_i)$ have AC-supports in $D(x_j)$. We say that a revision is {\em fruitful} if it deletes at least one value, while it is {\em redundant} if it achieves no pruning. 

%In the experiments presented in this paper we have used the {\em dom/wdeg} variable ordering heuristic which is one of the most efficient general-purpose heuristics for CSPs \cite{bhls04}. This heuristic assigns a weight to each constraint, initially set to one. Each time a constraint causes a domain wipeout (DWO) its weight is incremented by one. Each variable is associated with a {\em weighted degree} (wdeg), which is the sum of the weights over all constraints involving the variable and at least another (unassigned) variable. The {\em dom/wdeg} heuristic chooses the variable with minimum ratio of current domain size to weighted degree.

\subsection{maxRPC}

A value $a_i\in D(x_i)$  is {\em max restricted path consistent} (maxRPC) iff
it is AC and for each constraint $c_{ij}$ there exists a value $a_j\in
D(x_j)$ that is an AC-support of $a_i$ s.t. the pair of values
$(a_i,a_j)$ is {\em path consistent} (PC) \cite{db97}. A pair of values $(a_i,a_j)$ is PC iff for any third variable $x_k$ there exists a value $a_k\in D(x_k)$ s.t. $a_k$ is an AC-support of both $a_i$ and $a_j$. In this case $a_j$ is a {\em PC-support} of $a_i$ in $x_j$ and $a_k$ is a {\em PC-witness} for the pair $(a_i,a_j)$ in $x_k$. A variable is maxRPC iff all its values are maxRPC. A problem is maxRPC iff there is no empty domain and all variables are maxRPC.

To our knowledge, three algorithms for achieving maxRPC have been proposed in the literature so far. The first one, called \texttt{maxRPC1}, is a fine-grained
algorithm based on \texttt{AC6} and has optimal O($end^3$) time complexity and
O($end$) space complexity \cite{db97}. The second algorithm, called \texttt{maxRPC2}, is a coarse-grained algorithm having O($end^3$) time and O($ed$) space complexity \cite{grandoni03}. Finally, \texttt{maxRPC$^{rm}$} is a coarse-grained algorithm based on \texttt{AC3$^{rm}$} \cite{vion09}.
The time and space complexities of \texttt{maxRPC$^{rm}$} are O($en^2d^4$) and 
O($end$) but it has some advantages compared to the other two because of its lighter use of data structures. Among the three algorithms maxRPC2 seems to be the most promising for  stand-alone use as it has a better time and space complexity than \texttt{maxRPC$^{rm}$} without requiring heavy data structures or complex implementation as \texttt{maxRPC1} does. On the other hand, \texttt{maxRPC$^{rm}$} can be better suited for use during search as it avoids the costly maintainance of data structures.

Central to \texttt{maxRPC2} is the $LastPC$ data structure, as we call it here. For each constraint $c_{ij}$ and each value $a_i\in D(x_i)$, $LastPC_{x_i,a_i,x_j}$ gives the most recently discovered PC-support of $a_i$ in $D(x_j)$. \texttt{maxRPC2} maintains this data structure incrementally. This means that the data structure is copied when moving forward during search (i.e. after a successfully propagated variable assignment) and restored when backtracking (after a failed variable assignment). This results in the following behavior: When looking for a PC-support for $a_i$ in $D(x_j)$, \texttt{maxRPC2} first checks if $LastPC_{x_i,a_i,x_j}$ is valid. If it is not, it searches for a new PC-support starting from the value immediately after $LastPC_{x_i,a_i,x_j}$ in $D(x_j)$. In this way a good time complexity bound is achieved. On the other hand, \texttt{maxRPC$^{rm}$} uses a data structure similar to $LastPC$ to store {\em residues}, i.e. supports that have been discovered during execution and stored for future use, but does not maintain this structure incrementally\footnote{\texttt{maxRPC$^{rm}$} also uses residues in a different context.}. When looking for a PC-support for $a_i$ in $D(x_j)$, if the residue $LastPC_{x_i,a_i,x_j}$ is not valid then \texttt{maxRPC$^{rm}$} searches for a new PC-support from scratch in $D(x_j)$. This results in higher complexity, but crucially does not require costly maintainance of the $LastPC$ data structure during search. %Note that the relationship between \texttt{maxRPC2} and \texttt{maxRPC$^{rm}$} is similar to that between AC algorithms \texttt{AC2001/3.1} and \texttt{AC$^{rm}$}.  

A major overhead of both \texttt{maxRPC2} and \texttt{maxRPC$^{rm}$} is the following. When searching for a PC-witness for a pair of values $(a_i,a_j)$ in a third variable $x_k$, they always start the search from scratch, i.e. from the first available value in $D(x_k)$. As these searches can be repeated many times during search, there can be many redundant constraint checks. In contrast, \texttt{maxRPC1} manages to avoid searching from scratch through the use of an additional data structure. This saves many constraint checks, albeit resulting in O$(end)$ space complexity and requiring costly maintainance of this data structure during search. The algorithms we describe below largely eliminate these redundant constraint checks with lower space complexity, and in the case of \texttt{maxRPC3$^{rm}$} with only light use of data structures.

%In the following, we first describe ways to improve the performance of \texttt{maxRPC2} and \texttt{maxRPC$^{rm}$} combining ideas from the existing maxRPC algorithms as well as the AC algorithms \texttt{AC2001/3.1} and \texttt{AC$^{rm}$}. Then we present heuristics that can be used to boost the performance of any coarse-grained maxRPC algorithm. % by efficiently ordering certain operations. 

%Finally, we introduce an adaptive version of maxRPC that can be combined with any coarse-grained maxRPC algorithm.

\section{New Algorithms for maxRPC}
\label{sec:algorithms}

We first recall the basic ideas of algorithms \texttt{maxRPC2} and \texttt{maxRPC$^{rm}$} as described in \cite{grandoni03} and \cite{vion09}. Both algorithms use a propagation list $L$ where variables whose domain is pruned are added. Once a variable $x_j$ is removed from $L$ all neighboring variables are revised to delete any values that are no longer maxRPC. For any value $a_i$ of such a variable $x_i$ there are two possible reasons for deletion. The first, which we call {\em PC-support loss} hereafter, is when the unique PC-support $a_j\in D(x_j)$ for $a_i$ has been deleted. The second, which we call {\em PC-witness loss} hereafter, is when the unique PC-witness $a_j\in D(x_j)$ for the pair $(a_i, a_k)$, where $a_k$ is the unique PC-support for $a_i$ on some variable $x_k$, has been deleted. In both cases value $a_i$ is no longer maxRPC.

We now give a unified description of algorithms \texttt{maxRPC3} and \texttt{maxRPC3$^{rm}$}. Both algorithms utilize data structures $LastPC$ and $LastAC$ which have the following functionalities: For each constraint $c_{ij}\ $ and each value $\ a_i\in D(x_i)$, $LastPC_{x_i,a_i,x_j}\ $ and $LastAC_{x_i,a_i,x_j}$ give (point to) the most recently discovered PC and AC supports of $a_i$ in $D(x_j)$ respectively. Initially, all $LastPC$ and $LastAC$ pointers are set to a special value NIL, considered to precede all values in any domain. Algorithm \texttt{maxRPC3} updates the $LastPC$ and $LastAC$ structures incrementally like \texttt{maxRPC2} and \texttt{AC2001/3.1} respectively do. In contrast, algorithm \texttt{maxRPC3$^{rm}$} uses these structures as residues like \texttt{maxRPC$^{rm}$} and \texttt{AC$^{rm}$} do.

The pseudocode for the unified description of \texttt{maxRPC3} and \texttt{maxRPC3$^{rm}$} is given in Algorithm \ref{algMaxRPC3} and Functions~\ref{algRevise},~\ref{algsearchPCwit},~\ref{algWitness}. We assume the existence of a global Boolean variable RM which determines whether the algorithm presented is instantiated to \texttt{maxRPC3} or to \texttt{maxRPC3$^{rm}$}. If RM is true, the algorithm used is \texttt{maxRPC3$^{rm}$}. Otherwise, the algorithm is \texttt{maxRPC3}. 

Being coarse-grained, Algorithm \ref{algMaxRPC3} uses a propagation list $L$ where variables that have their domain filtered are inserted. If the algorithm is used for preprocessing then, during an initialization phase, for each value $a_i$ of each variable $x_i$ we check if $a_i$ is maxRPC. If it is not then it is deleted from $D(x_i)$ and $x_i$ is added to $L$. The initialization function is not shown in detail due to limited space. If the algorithm is used during search then $L$ is initialized with the currently assigned variable (line 3).

In the main part of Algorithm~\ref{algMaxRPC3}, when a variable $x_j$ is removed from $L$, each variable $x_i$ constrained with $x_j$ must be made maxRPC. For each value $a_i\in D(x_i)$ Algorithm \ref{algMaxRPC3}, like \texttt{maxRPC2} and \texttt{maxRPC$^{rm}$}, establishes if $a_i$ is maxRPC by checking for PC-support loss and PC-witness loss at lines 8 and 12. 

\begin{algorithm}
\centering
\begin{scriptsize}
\caption{\texttt{maxRPC3/maxRPC3$^{rm}$}}
\label{algMaxRPC3}
\begin{algorithmic}[1]
\IF {$\lnot$ RM}
	%\IF {$\lnot${\em initialization}(L, LastPC, LastAC)} 
	%	\STATE return FAILURE;
	%\ENDIF
	\STATE {\bf if} $\lnot${\em initialization}(L, LastPC, LastAC) {\bf then} return 
																									FAILURE;
\ENDIF
%\ELSE 
\STATE {\bf else} L = $\{$currently assigned variable$\}$;
%\ENDIF
\WHILE {L $\neq \O$}
	\STATE L=L$-\{x_j\}$;
	\FOR {\textbf{each} $x_i \in X$ s.t. $c_{ij}\in C$}
		\FOR {\textbf{each} $a_i \in D(x_i)$}
			\IF {$\lnot${\em searchPCsup}$(a_i,x_j)$}
				\STATE delete $a_i$;
				\STATE L=L $ \cup\ \{x_i\}$;
			\ELSE
				\IF {$\lnot${\em checkPCwit}$(a_i,x_j)$}
					\STATE delete $a_i$;
					\STATE L=L $ \cup\ \{x_i\}$;
				\ENDIF
			\ENDIF
		\ENDFOR

	%\IF {$D(x_i)$ is empty} 
		\STATE {\bf if} $D(x_i)$ is empty {\bf then} return FAILURE;
	%\ENDIF
	\ENDFOR

\ENDWHILE

\STATE return SUCCESS;

\end{algorithmic}
\end{scriptsize}
\end{algorithm}

First, function {\em searchPCsup} is called to check if a PC-support for $a_i$ exists in $D(x_j)$. If value $LastPC_{x_i,a_i,x_j}$ is still in $D(x_j)$, then {\em searchPCsup} returns TRUE (lines 1-2). If $LastPC_{x_i,a_i,x_j}$ is not valid, we search for a new PC-support. If \texttt{maxRPC3} is used, we can take advantage of the $LastPC$ and $LastAC$ pointers to avoid starting this search from scratch. Specifically, we know that no PC-support can exist before $LastPC_{x_i,a_i,x_j}$, and also none can exist before $LastAC_{x_i,a_i,x_j}$, since all values before $LastAC_{x_i,a_i,x_j}$ are not AC-supports of $a_i$. Lines 5-6 in  {\em searchPCsup} take advantage of these to locate the appropriate starting value $b_j$. Note that \texttt{maxRPC2} always starts the search for a PC-support from the value after $LastPC_{x_i,a_i,x_j}$. If the algorithm is called during search, in which case we use \texttt{maxRPC3$^{rm}$} then the search for a new PC-support starts from scratch (line 8), just like \texttt{maxRPC$^{rm}$} does.

For every value $a_j\in D(x_j)$, starting with $b_j$, we first check if is an AC-support of $a_i$ (line 10). This is done using function {\em isConsistent} which simple checks if two values are compatible. If it is, and the algorithm is \texttt{maxRPC3}, then we can update $LastAC_{x_i,a_i,x_j}$ under a certain condition (lines 12-13). Specifically, if $LastAC_{x_i,a_i,x_j}$ was deleted from $D(x_j)$, then we can set $LastAC_{x_i,a_i,x_j}$ to $a_j$ in case $LastAC_{x_i,a_i,x_j}>LastPC_{x_i,a_i,x_j}$. If $LastAC_{x_i,a_i,x_j} \leq LastPC_{x_i,a_i,x_j}$ then we cannot do this as there may be AC-supports for $a_i$ between $LastAC_{x_i,a_i,x_j}$ and $LastPC_{x_i,a_i,x_j}$ in the lexicographical ordering. We then move on to verify the path consistency of $(a_i,a_j)$ through function {\em searchPCwit}. 

If no PC-support for $a_i$ is found in $D(x_j)$, {\em searchPCsup} will return FALSE, %which means that 
$a_i$ will be deleted and $x_i$ will be added to $L$. Otherwise, $LastPC_{x_i,a_i,x_j}$ is set to the discovered PC-support $a_j$ (line 15). If \texttt{maxRPC3$^{rm}$} is used then we update the residue $LastAC_{x_i,a_i,x_j}$ since the discovered PC-support is also an AC-support. 
In addition, to exploit the multidirectionality of residues, \texttt{maxRPC3$^{rm}$} sets $LastPC_{x_j,a_j,x_i}$ to $a_i$, as in \cite{vion09}.

\begin{algorithm}
\centering
\begin{scriptsize}
\floatname{algorithm}{Function}
\caption{{\em searchPCsup}($a_i$, $x_j$):\textbf{boolean}}
\label{algRevise}
\begin{algorithmic}[1]

\IF {$LastPC_{x_i,a_i,x_j} \in D(x_j)$} 
	\STATE return \TRUE;
\ELSE
	\IF {$\lnot$ RM}
		%\IF {$LastAC_{x_i,a_i,x_j} \in D(x_j)$} 
		\STATE {\bf if} $LastAC_{x_i,a_i,x_j} \in D(x_j)$ {\bf then} 
			%\STATE 
			$b_j$ = max($LastPC_{x_i,a_i,x_j}$+1,$LastAC_{x_i,a_i,x_j}$);
		%\ELSE
			\STATE {\bf else} $b_j$ = 
					max($LastPC_{x_i,a_i,x_j}$+1,$LastAC_{x_i,a_i,x_j}$+1);
		%\ENDIF
		\ELSE \STATE $b_j$ = first value in $D(x_j)$;
	\ENDIF
	\FOR {\textbf{each} $a_j \in D(x_j)$, $a_j \geq b_j$ }
		\IF {{\em isConsistent}$(a_i,a_j)$}
			\IF {$\lnot RM$}
				\IF {$LastAC_{x_i,a_i,x_j} \notin D(x_j)$ AND $LastAC_{x_i,a_i,x_j}>LastPC_{x_i,a_i,x_j}$}
					\STATE $LastAC_{x_i,a_i,x_j}=a_j$;
				\ENDIF
			\ENDIF
			\IF {{\em searchPCwit}$(a_i,a_j)$}
				\STATE $LastPC_{x_i,a_i,x_j} = a_j$;
				\STATE {\bf if} RM {\bf then} 
				%\IF {$\lnot$PREPROCESSING}
				%\STATE 
				$LastAC_{x_i,a_i,x_j} = a_j$; $LastPC_{x_j,a_j,x_i} = a_i$;
				%\ENDIF
				\STATE return \TRUE;
			\ENDIF
		\ENDIF
	\ENDFOR
\ENDIF

\STATE return \FALSE;

\end{algorithmic}
\end{scriptsize}
\end{algorithm}

Function {\em searchPCwit} checks if a pair of values ($a_i$,$a_j$) is PC by doing the following for each variable $x_k$ constrained with $x_i$ and $x_j$\footnote{Since AC is enforced by the maxRPC algorithm, we only need to consider variables that form a 3-clique with $x_i$ and $x_j$.}. 
First, it checks if either $LastAC_{x_i,a_i,x_k}$ is valid and consistent with $a_j$ or $LastAC_{x_j,a_j,x_k}$ is valid and consistent with $a_i$ (line 3). If one of these conditions holds then we have found a PC-witness for ($a_i$,$a_j$) without searching in $D(x_k)$ and we move on to the next variable constrained with $x_i$ and $x_j$. Note that neither \texttt{maxRPC2} nor \texttt{maxRPC$^{rm}$} can do this as they do not have the $LastAC$ structure. Experimental results in Section~\ref{sec:experiments} demonstrate that these simple conditions can eliminate a very large number of redundant constraint checks. 

If none of the conditions holds then we have to search in $D(x_k)$ for a PC-witness. If the algorithm is \texttt{maxRPC3} then we can exploit the $LastAC$ structure to start this search from $b_k=max\{LastAC_{x_i,a_i,x_k}, LastAC_{x_j,a_j,x_k}\}$ (line 6). But before doing this, we call function {\em seekACsupport} (not shown for space reasons), first with $(x_i,a_i,x_k)$ and then with $(x_j,a_j,x_k)$ as parameters, to find the lexicographically smallest AC-supports for $a_i$ and $a_j$ in $D(x_k)$ (line 5). If such supports are found, $LastAC_{x_i,a_i,x_k}$ and $LastAC_{x_j,a_j,x_k}$ are updated accordingly. In case no AC-support is found for either $a_i$ or $a_j$ then {\em seekACsupport} returns FALSE, and subsequently searchPCwit() will also return FALSE.

\begin{algorithm}%[H]
\centering
\begin{scriptsize}
\floatname{algorithm}{Function}
\caption{{\em searchPCwit}$(a_i,a_j)$:\textbf{boolean}}
\label{algsearchPCwit}
\begin{algorithmic}[1]

\FOR {\textbf{each} $x_k \in V$ s.t. $c_{ik} \in C$ and  $c_{jk} \in C$}
	\STATE maxRPCsupport=FALSE;
	%\IF 
	\STATE{\bf if} {($LastAC_{x_i,a_i,x_k}\in D(x_k)$ AND {\em isConsistent}$(LastAC_{x_i,a_i,x_k},a_j)$) OR ($LastAC_{x_j,a_j,x_k}\in D(x_k)$ AND {\em isConsistent}$(LastAC_{x_j,a_j,x_k},a_i)$)} {\bf then continue};
		%\STATE {\bf continue};
	%\ENDIF
	\IF {$\lnot$ RM}
		\STATE {\bf if} $\lnot seekACsupport(x_i,a_i,x_k)$ OR $\lnot seekACsupport(x_j,a_j,x_k)$ {\bf then}
				return {\bf false};
		\STATE $b_k = max(LastAC_{x_i,a_i,x_k}, LastAC_{x_j,a_j,x_k})$;
	\ENDIF
	%\ELSE 
	%	\STATE $b_k$ = first value in $D(x_k)$; 
	%\STATE {\bf if} PREPROCESSING {\bf then} 
						%$b_k = max(LastAC_{x_i,a_i,x_k}+1, LastAC_{x_j,a_j,x_k}+1)$;
	\STATE {\bf else} $b_k$ = first value in $D(x_k)$; 	
	%\ENDIF
	\FOR {\textbf{each} $a_k \in D(x_k)$, $a_k\geq b_k$}
			%IF {PREPROCESSING}
				%\STATE {\bf if} {$LastAC_{x_i,a_i,x_k} \notin D(x_k)$ AND {\em isConsistent}$(a_i,a_k)$} {\bf then} 				
				%							$LastAC_{x_i,a_i,x_k} = a_k$;
				%\STATE {\bf if} {$LastAC_{x_j,a_j,x_k} \notin D(x_k)$ AND {\em isConsistent}$(a_j,a_k)$} {\bf then} 
				%							$LastAC_{x_j,a_j,x_k} = a_k$;
			%\ENDIF
			\IF {{\em isConsistent}$(a_i,a_k)$ AND {\em isConsistent}$(a_j,a_k)$}
				%\IF {$\lnot$PREPROCESSING}
				%	\STATE $LastAC_{x_i,a_i,x_k} = LastAC_{x_j,a_j,x_k} = a_k$;
				%\ENDIF
				\STATE {\bf if} RM {\bf then} 
							$LastAC_{x_i,a_i,x_k} = LastAC_{x_j,a_j,x_k} = a_k$; 
				%\STATE $LastAC_{x_i,a_i,x_k} = LastAC_{x_j,a_j,x_k} = a_k$;
				\STATE maxRPCsupport=TRUE; {\bf break};
				%\STATE break;
			\ENDIF
	\ENDFOR

	\STATE {\bf if} $\lnot$maxRPCsupport {\bf then} return \FALSE;
	%\IF {$\lnot$maxRPCsupport}
	%	\STATE return \FALSE;
	%\ENDIF
\ENDFOR

\STATE return \TRUE;

\end{algorithmic}
\end{scriptsize}
\end{algorithm}

If the algorithm used is \texttt{maxRPC3$^{rm}$} then we start search for a PC-witness from scratch (line 7), as \texttt{maxRPC2} and \texttt{maxRPC$^{rm}$} always do. If a PC-witness $a_k$ is found (line 9) and we are using \texttt{maxRPC3$^{rm}$} then both residues $LastAC_{x_i,a_i,x_k}$ and $LastAC_{x_j,a_j,x_k}$ are set to $a_k$ as they are the most recently discovered AC-supports. If no PC-witness is found then we have determined that the pair ($a_i$,$a_j$) is not PC and as a result FALSE will be returned and {\em searchPCsup} will move to check if the next available value in $D(x_j)$ is a PC-support for $a_i$. 

If value $a_i$ is not removed by {\em searchPCsup} in Algorithm \ref{algMaxRPC3}, {\em checkPCwit} is called to check for PC-witness loss. This is done by iterating over the variables that are constrained with both $x_i$ and $x_j$. For each such variable $x_k$, we first check if  $a_k=LastPC_{x_i,a_i,x_k}$ is still in $D(x_k)$ (line 3). If so then we check if there still is a PC-witness in $D(x_j)$. This is done by first checking if either $LastAC_{x_i,a_i,x_j}$ is valid and consistent with $a_k$ or $LastAC_{x_k,a_k,x_j}$ is valid and consistent with $a_i$ (line 4). If neither of these conditions holds then we search for a PC-witness starting from $b_j=max\{LastAC_{x_i,a_i,x_j}, LastAC_{x_k,a_k,x_j}\}$ in case of \texttt{maxRPC3} (line 9), after checking the existence of AC-supports for $a_i$ and $a_k$ in $D(x_j)$, by calling {\em seekACsupport} (line 8). If there is no AC-support in $D(x_j)$ for either $a_i$ or $a_k$ we set the auxiliary Boolean variable {\em findPCsupport} to TRUE to avoid searching for a PC-witness. 

If \texttt{maxRPC3$^{rm}$} is used, we start searching for a PC-witness from scratch (line 11). Note that \texttt{maxRPC2} does not do the check of line 4 and always starts the search for a PC-witness from the first value in $D(x_j)$. In contrast, \texttt{maxRPC$^{rm}$} avoids some redundant checks through the use of special residues, albeit resulting in O$(end)$ space complexity.
When using \texttt{maxRPC3$^{rm}$}, for each value $a_j\in D(x_j)$ we check if it is compatible with $a_i$ and $a_k$ and move the $LastAC$ pointers accordingly (lines 14-15), exploiting the multidirectionality of residues,

\begin{algorithm}%[H]
\centering
\begin{scriptsize}
\floatname{algorithm}{Function}
\caption{{\em checkPCwit}$(a_i,x_j)$:\textbf{boolean}}
\label{algWitness}
\begin{algorithmic}[1]

\FOR {\textbf{each} $x_k \in V$ s.t. $c_{ik}\in C$ and $c_{kj} \in C$}
	\STATE witness=FALSE; findPCsupport=FALSE;
	\IF {$a_k = LastPC_{x_i,a_i,x_k} \in D(x_k)$}
		\IF {($LastAC_{x_i,a_i,x_j} \in D(x_j)$ AND {\em isConsistent}$(LastAC_{x_i,a_i,x_j},a_k)$) OR ($LastAC_{x_k,a_k,x_j} \in D(x_j)$ AND {\em isConsistent}$(LastAC_{x_k,a_k,x_j},a_i)$)}
			\STATE witness=TRUE;
		\ELSE 
			\IF {$\lnot$ RM}
				\IF {$seekACsupport(x_i,a_i,x_j)$ AND $seekACsupport(x_k,a_k,x_j)$}
					\STATE $b_j=max(LastAC_{x_i,a_i,x_j}, LastAC_{x_k,a_k,x_j})$;
				\ENDIF
				\STATE {\bf else} findPCsupport=TRUE;  
			\ENDIF
			%\STATE {\bf if} PREPROCESSING {\bf then} 
			%					$b_j = max(LastAC_{x_i,a_i,x_j}+1, LastAC_{x_k,a_k,x_j}+1)$;
			\STATE {\bf else} $b_j$ = first value in $D(x_j)$; 
			\IF {$\lnot findPCsupport$}
				\FOR {\textbf{each} $a_j \in D(x_j)$, $a_j\geq b_j $}
						\IF {{\em isConsistent}$(a_i,a_j)$ AND {\em isConsistent}$(a_k,a_j)$}
							%\IF {$\lnot$PREPROCESSING}
							%	\STATE $LastAC_{x_i,a_i,x_j} = LastAC_{x_k,a_k,x_j} = a_j$;
							%\ENDIF
							\STATE {\bf if} RM {\bf then}
									$LastAC_{x_i,a_i,x_j} = LastAC_{x_k,a_k,x_j} = a_j$;
							\STATE witness=TRUE; {\bf break};
							%\STATE break;
						\ENDIF
				\ENDFOR
			\ENDIF
		\ENDIF
	\ENDIF

	\IF{$\lnot$witness AND exists $a_k>LastPC_{x_i,a_i,x_k} \in D(x_k)$}
	\IF {$\lnot$ RM}
		%\IF {$LastAC_{x_i,a_i,x_k} \in D(x_k)$} 
		%	\STATE $b_k$ = max($LastPC_{x_i,a_i,x_k}$+1,$LastAC_{x_i,a_i,x_k}$);
		%\ELSE
		%	\STATE $b_k$ = max($LastPC_{x_i,a_i,x_k}$+1,$LastAC_{x_i,a_i,x_k}$+1);
		%\ENDIF
		\STATE {\bf if} $LastAC_{x_i,a_i,x_k} \in D(x_k)$ {\bf then}
					 $b_k$ = max($LastPC_{x_i,a_i,x_k}$+1,$LastAC_{x_i,a_i,x_k}$);
		\STATE {\bf else} 
					 $b_k$ = max($LastPC_{x_i,a_i,x_k}$+1,$LastAC_{x_i,a_i,x_k}$+1
	\ELSE	 \STATE $b_k$ = first value in $D(x_k)$;
	\ENDIF	
		\FOR {\textbf{each} $a_k \in D(x_k)$, $a_k\geq$ $b_k$}
			\IF {{\em isConsistent}$(a_i,a_k)$}
				\IF {$\lnot$ RM}
					\IF {$LastAC_{x_i,a_i,x_k} \notin D(x_k)$ AND $LastAC_{x_i,a_i,x_k}>LastPC_{x_i,a_i,x_k}$}
						\STATE $LastAC_{x_i,a_i,x_k}=a_k$;
					\ENDIF
				\ENDIF
				\IF {{\em searchPCwit}$(a_i,a_k)$}
					\STATE $LastPC_{x_i,a_i,x_k}=a_k$;
					%\IF {$\lnot$PREPROCESSING} 
					%	\STATE $LastAC_{a_i,x_k}=a_k$;
					%\ENDIF
					\STATE {\bf if} RM {\bf then} 
							$LastAC_{x_i,a_i,x_k}=a_k$; $LastPC_{x_k,a_k,x_i}=a_i$;
					\STATE witness=TRUE; {\bf break};
					%\STATE {\em break};
				\ENDIF
			\ENDIF
		\ENDFOR
	\ENDIF

%\IF {$\lnot$witness}
%	\STATE return \FALSE;
%\ENDIF
\STATE {\bf if} $\lnot$witness {\bf then} return \FALSE;
\ENDFOR

\STATE return \TRUE;

\end{algorithmic}
\end{scriptsize}
\end{algorithm}

If $LastPC_{x_i,a_i,x_k}$ has been removed or $a_i$ has no PC-witness in $D(x_j)$, we search for a new PC-support for $a_i$ in $D(x_k)$. As in function {\em searchPCsup}, when \texttt{maxRPC3} is used this search starts at an appropriate value calculated taking advantage of $LastPC_{x_i,a_i,x_k}$ and  $LastAC_{x_i,a_i,x_k}$ (lines 18-20). When \texttt{maxRPC3$^{rm}$} is used we start from scratch. If an AC-support for $a_i$ is found (line 24), we check if it is also a PC-support by calling function {\em searchPCwit} (line 28). If \texttt{maxRPC3} is used then $LastAC_{x_i,a_i,x_k}$ is updated when necessary (lines 26-27). If a PC-support is found, $LastPC_{x_i,a_i,x_k}$ is set accordingly (line 29). If \texttt{maxRPC3$^{rm}$} is used then the residue $LastAC_{x_i,a_i,x_k}$ is also updated, as is $LastPC_{x_k,a_k,x_i}$ (bidirectionality). If the search for a PC-support fails then FALSE will be returned, $a_i$ will be deleted, and $x_i$ will be added to L.

\subsection{Light maxRPC}

\label{subsection:light}

\vspace{-2mm}

Light maxRPC (lmaxRPC) is an approximation of maxRPC that only propagates the loss of AC-supports and not the loss of PC-witnesses \cite{vion09}. %That is, when removing a variable $x_j$ from $L$, for each $a_i \in D(x_i)$, where $x_i$ is constrained with $x_j$, lmaxRPC only checks if there is a PC-support of $a_i$ in $D(x_j)$. 
This ensures that the obtained algorithm enforces a consistency property that is at least as strong as AC. 

lmaxRPC is a procedurally defined local consistency, meaning that its description is tied to a specific maxRPC algorithm. Light versions of algorithms \texttt{maxRPC3} and \texttt{maxRPC3$^{rm}$}, simply noted \texttt{lmaxRPC3} and \texttt{lmaxRPC3$^{rm}$} respectively, can be obtained by omitting the call to the {\em checkPCwit} function (lines 11-14 of Algorithm \ref{algMaxRPC3}). %Also, to exploit the bidirectionality of residues as in \cite{vion09}, the highlighted assignment $LastPC_{x_j,a_j,x_i} = a_i$ in line 16 of {\em searchPCsup} is added. Bidirectionality simply means that if $a_j$ is a PC-support for $a_i$ then so is $a_i$ for $a_j$.
In a similar way, we can obtain light versions of algorithms \texttt{maxRPC2} and \texttt{maxRPC$^{rm}$}.

As already noted in \cite{vion09}, the light versions of different maxRPC algorithms may not be equivalent in terms of the pruning they achieve. To give an example, a brute force algorithm for lmaxRPC that does not use any data structures can achieve more pruning than algorithms \texttt{lmaxRPC2}, \texttt{lmaxRPC3}, and \texttt{lmaxRPC}$^{rm}$, albeit being much slower in practice. %This is because in {\em searchPCsup} such an algorithm will always search for a PC-support in the domain of $x_j$ from scratch. 
Consider that any of these three algorithms will return TRUE in case $LastPC_{x_i,a_i,x_j}$ is valid. However, although $LastPC_{x_i,a_i,x_j}$ is valid, it may no longer be a PC-support because the PC-witness in some third variable may have been deleted, and it may be the last one. In a case where $LastPC_{x_i,a_i,x_j}$ was the last PC-support in $x_j$ for value $a_i$, the three advanced algorithms will not delete $a_i$ while the brute force one will. This is because it will exhaustively check all values of $x_j$ for PC-support, concluding that there is none.

%If we also omit the highlighted code in Algorithm \texttt{maxRPC3} and Functions {\em searchPCsup} and {\em searchPCwit} then we get \texttt{lmaxRPC2}, the light version of \texttt{maxRPC2}. In turn, algorithm \texttt{lmaxRPC}$^{rm}$ can be obtained from \texttt{lmaxRPC2} if we simply remove condition ``$a_j \geq b_j$'' in line 9 of Function {\em searchPCsup}. In this case, the value pointed to by $LastPC_{x_i,a_i,x_j}$ only acts as a residue \cite{vion09}. If it has been deleted, search for a new PC-support starts again from the first value in $D(x_j)$. Finally, if we also remove lines 1-3 from {\em searchPCsup} and any other reference to the $LastPC$ structure then we get a brute-force lmaxRPC algorithm that does not employ any data structures at all.

The worst-case time and space complexities of algorithm \texttt{lmaxRPC2} are the same as maxRPC2. Algorithm \texttt{lmaxRPC}$^{rm}$ has O$(n^3d^4)$ time and O$(ed)$ space complexities, which are lower than those of \texttt{maxRPC}$^{rm}$. Experiments with random problems using algorithms \texttt{lmaxRPC}$^{rm}$ and \texttt{maxRPC}$^{rm}$ showed that the pruning power of lmaxRPC is only slightly weaker than that of maxRPC \cite{vion09}. At the same time, it can offer significant gains in run times when used during search. These results %The run time benefits of an algorithm that applies lmaxRPC during search, as opposed to maxRPC, 
were also verified by us through a series of experiments on various problem classes.

\vspace{-3mm}

\subsection{Correctness and Complexities}
%\vspace{-2mm}

We now prove the correctness of algorithms \texttt{maxRPC3} and \texttt{maxRPC3$^{rm}$} and analyze their worst-case time and space complexities.

\begin{proposition}\rm
Algorithm \texttt{maxRPC3} is sound and complete.
\newline 
\\
\begin{proof}\textbf{\textit{Soundness.}}
To prove the soundness of \texttt{maxRPC3} we must prove that any value that is deleted by \texttt{maxRPC3} is not maxRPC. %The algorithm can delete a value only when, having removed a variable $x_j$ from $Q$, it examines the variables constrained with $x_j$ to remove any inconsistent values from their domains. 
Let $a_i \in D(x_i)$ be a value that is deleted by \texttt{maxRPC3}. It is either removed from $D(x_i)$ during the initialization phase (line 15) or in line 8 of Algorithm \ref{algMaxRPC3}, after \textit{searchPCsup} has returned \textit{false}, or in line 12, after \textit{searchPCsup} has returned \textit{true} and \textit{checkPCwit} has returned \textit{false}. 

In the first case, since function \textit{initilization} checks all values in a brute-force manner, it is clear that any deleted value $a_i$ either has no AC-support or none of its AC-supports is a PC-support in some variable $x_j$. The non-existence of a PC-support is determined using function \textit{searchPCwit} whose correctness is discussed below.

In the second case, since \textit{searchPCsup} returns \textit{false}, $LastPC_{x_i,a_i,x_j}$ is not valid so a new PC-support in $D(x_j)$ is seeked (lines 9-17). This search starts with the value at max($LastPC_{x_i,a_i,x_j}$+1, $LastAC_{x_i,a_i,x_j}$) or at max($LastPC_{x_i,a_i,x_j}$ +1, $LastAC_{x_i,a_i,x_j}$ +1), depending on whether $LastAC_{x_i,a_i,x_j}$ is valid or not. This is correct since any value before $LastPC_{x_i,a_i,x_j}$+1 and any value before $LastAC_{x_i,a_i,x_j}$ is definitely not an AC-support for $a_i$ (similarly for the other case). \textit{searchPCsup} will return false either because no AC-support for $a_i$ can be found in $D(x_j)$ (line 10), or because for any AC-support found, \textit{searchPCwit} returned false (line 13). In the former case there is no PC-support for $a_i$ in $D(x_j)$ since there is no AC-support. In the latter case, for any AC-support $a_j$ found there must be some third variable $x_k$ for which no PC-witness for the pair $(a_i,a_j)$ exists. For each third variable $x_k$ \textit{searchPCwit} correctly identifies a PC-witness if one of the conditions in line 3 holds. In none holds then \textit{searchPCwit} searches for a PC-witness starting from max($LastAC_{x_i,a_i,x_k}$, $LastAC_{x_j,a_j,x_k}$) (line 6). This is correct since $LastAC_{x_i,a_i,x_k}$ and  $LastAC_{x_j,a_j,x_k}$ are updated with the lexicographically smallest support of $a_i$ (resp. $a_j$) in $D(x_k)$ by calling function \textit{seekACsup}, meaning that any value smaller than max($LastAC_{x_i,a_i,x_k}$, $LastAC_{x_j,a_j,x_k}$) is incompatible with either $a_i$ or $a_j$. Therefore, if \textit{searchPCwit} returns \textit{false} then there is no PC-witness for some third variable $x_k$. Hence, if \textit{searchPCsup} returns \textit{false}, it means no PC-support for $a_i$ can be found in $D(x_j)$ and it is thus correctly deleted. 

Now assume that the call to \textit{searchPCsup} returned \textit{true} and $a_i$ was removed after \textit{checkPCwit} returned \textit{false}. This means that for some variable $x_k$, constrained with both $x_i$ and $x_j$, both the first part (lines 3-11) and the second part (lines 13-24) of  \textit{checkPCwit} failed to set the Boolean {\em witness} to \textit{true}. Regarding the first part, the failure means that the pair of values $(a_i,a_k)$, where $a_k$ is the last PC-support of $a_i$ in $D(x_k)$ found, has no PC-witness in $D(x_j)$. In more detail, the search for a PC-witness correctly starts from max($LastAC_{x_i,a_i,x_j}$, $LastAC_{x_j,a_j,x_j}$) in line 9, after both $LastAC$ pointers have been updated by \textit{seekACsup}. The condition in line 4 is similar to the corresponding condition in \textit{searchPCwit} and thus, if it is true, the search for PC-witness is correctly overriden. Regarding the second part, the failure means that no alternative PC-support for $a_i$ in $D(x_k)$ was found. In more detail, the search for a PC-support starts from max($LastPC_{x_i,a_i,x_k}$+1, $LastAC_{x_i,a_i,x_k}$) or max($LastPC_{x_i,a_i,x_k}$+1, $LastAC_{x_i,a_i,x_k}$+1), depending on the existence of $LastAC_{x_i,a_i,x_k}$. This is correct since no ealier value can be a PC-support. If there is no consistent ($a_i,a_k$) pair or \textit{searchPCwit} returns \textit{false} for all consistent pairs found, then $a_i$ has no PC-support in $D(x_k)$ and is thus correctly deleted. 

\textbf{\textit{Completeness.}}
To prove the completeness of \texttt{maxRPC3} we need to show that if a value is not maxRPC then the algorithm will delete it. The initialization function checks all values of all variables one by one in a brute-force manner and removes any value that is not maxRPC. Values that are maxRPC have their $LastPC$ pointers set to the discovered PC-supports. Thereafter, the effects of such removals are propagated by calling Algorithm \ref{algMaxRPC3} and as a result new value deletions may occur. Now consider a value $a_i \in D(x_i)$ that was not removed by the initialization function but after propagation is no longer maxRPC. This is either because of PC-support or PC-witness loss. 

In the first case assume that $x_j$ is the variable in which $a_i$ no longer has a PC-support. Since the previously found PC-support of $a_i$ has been deleted, $x_j$ must have been added to $Q$ at some point. When $x_j$ is removed from $Q$ all neighboring variables, including $x_i$ will be checked. Function \textit{searchPCsup} will find that $LastPC_{x_i,a_i,x_j}$ is no longer valid and will search for a new PC-support concluding that there is none. Therefore, it will return \textit{false} and $a_i$ will be deleted.  

In the second case assume that the pair of values ($a_i$,$a_j$), where $a_j$ is the last PC-support of $a_i$ in $D(x_j)$, has lost its last PC-witness $a_k$ in variable $x_k$. If $LastPC_{x_i,a_i,x_j}$ is not valid, which means that $x_j$ was added to $Q$, then we have the same case as above. Therefore, after $x_j$ is removed from $Q$, \textit{searchPCsup} will find out that there is no PC-support for $a_i$ in $D(x_j)$ and will delete it. If $LastPC_{x_i,a_i,x_j}$ is valid then \textit{searchPCsup} will return \textit{true} (line 2). Since $a_k$ was deleted, $x_k$ was added to $Q$ at some point.  When $x_k$ is removed from $Q$ all neighboring variables, including $x_i$ will be checked. If $a_i$ has no longer a PC-support in $D(x_k)$, this will be detected by \textit{searchPCsup} and $a_i$ will be deleted. Otherwise, function \textit{checkPCwit} will be called. The for loop in line 1 will go through every variable constrained with both $x_i$ and $x_k$, including $x_j$. Since $LastPC_{x_i,a_i,x_j}$ is valid, a new PC-witness for ($a_i$,$a_j$) in $D(x_k)$ will be seeked (lines 3-11). Since $a_k$ was the last PC-witness, none will be found and as a result a new PC-support for $a_i$ in $D(x_j)$ will be seeked (lines 13-24). Since $a_j$ was the last PC-support for $a_i$ in $D(x_j)$, none will be found, \textit{checkPCwit} will return \textit{false}, and $a_i$ will be deleted.
\end{proof}
\end{proposition}

\begin{proposition}\rm
Algorithm \texttt{maxRPC3$^{rm}$} is sound and complete.
\newline 
\\
\begin{proof}The proof is very similar to the corresponding proof for \texttt{maxRPC3}. As explained, the main difference between the two algorithms concerns the use of the $LastAC$ and $LastPC$ structures. As \texttt{maxRPC3$^{rm}$} does not maintain these structures incrementally, the searches for PC-supports in \textit{searchPCsup} and \textit{checkPCwit} and the searches for PC-witnesses in \textit{searchPCwit} and \textit{checkPCwit} start from scratch. Clearly, this has no effect on the soundness or completeness of the algorithm since it guarantees that all potential PC-supports and PC-witnesses are checked. Furthermore, the conditions for avoiding redundant searches using residues are the same as in \texttt{maxRPC3}. Finally, another difference between the two algorithms is the exploitation of bidirectionality by \texttt{maxRPC3$^{rm}$}. By the definition of path and arc consistency, bidirectionality holds. That is, when a PC-support (AC-support) $a_j \in D(x_j)$ is located for a value $a_i \in D(x_i)$ then $a_i$ is a PC-support (AC-support) for $a_j$. Since the property of bidirectionality is exploited only to update residues, it does not affect the correctness of the algorithm.
\end{proof}
\end{proposition}

We now discuss the complexities of algorithms \texttt{maxRPC3} and \texttt{maxRPC3$^{rm}$} and their light$\ $ versions. To directly $\ $compare with existing$\ $ algorithms$\ $ for (l)maxRPC, the time complexities give the asymptotic number of constraint checks\footnote{However, constraint checks do not always reflect run times as other operations may have an equal or even greater effect.}. Folllowing \cite{liki07}, the {\em node} time (resp. space) complexity of a (l)maxRPC algorithm is the worst-case time (resp. space) complexity of invoking the algorithm after a variable assignment.  The corresponding {\em branch} complexities of an (l)maxRPC algorithm are the worst-case complexities of any incremental sequence of $k\leq n$ invocations of the algorithm. That is, the complexities of incrementally running the algorithm down a branch of the search tree until a fail occurs. 

\begin{proposition}\rm
The node and branch time complexity of \texttt{(l)maxRPC3} is O$(end^3)$.
\newline 
\\
\begin{proof}The complexity is determined by the total number of calls to function {\em isConsistent} in {\em searchPCsup}, {\em checkPCwit}, and mainly {\em searchPCwit} where most checks are executed.

Each variable can be inserted and extracted from $L$ every time a value is deleted from its domain, giving O$(d)$ times in the worst case. Each time a variable $x_j$ is extracted from $L$, {\em searchPCsup} will look for a PC-support in $D(x_j)$ for all values $a_i \in D(x_i)$, s.t. $c_{i,j} \in C$. For each variable $x_i$, O$(d)$ values are checked. Checking if a value $a_j\in D(x_j)$ is a PC-support involves first checking in O$(1)$ if it is an AC-support (line 9 in \textit{searchPCsup}) and then calling {\em searchPCwit}. The cost of {\em searchPCwit} is O$(n + nd)$ since there are O$(n)$ variables constrained with both $x_i$ and $x_j$ and, after making the checks in line 3, their domains must be searched for a PC-witness, each time from scratch with cost O$(nd)$. Through the use of $LastPC$ no value of $x_j$ will be checked more than once over all the O$(d)$ times $x_j$ is extracted from $L$, meaning that for any value $a_i\in D(x_i)$ and any variable $x_j$, the overall cost of {\em searchPCwit} will be O$(dn + nd^2)$ = O$(nd^2)$. Hence, {\em searchPCsup} will cost O$(nd^2)$ for one value of $x_i$, giving O$(nd^3)$ for $d$ values. Since, in the worst case, this process will be repeated for every pair of variables $x_i$ and $x_j$ that are constrained, 
%(i.e. for every constraint), 
the total cost of {\em searchPCsup} will be O$(end^3)$. This is the node complexity of \texttt{lmaxRPC3}.

In {\em checkPCwit} the algorithms iterate over the variables in a triangle with $x_j$ and $x_i$. %The number of times that the outer {\bf for} loop of {\em checkPCwit} is executed (for one value $a_i\in D(x_i)$) will be O$(e)$, since the maximum number of pairs of variables that are constrained with $x_j$ and with each other is bounded by $e$. 
In the worst case, for each such variable $x_k$, $D(x_j)$ will be searched from scratch for a PC-witness of $a_i$ and its current PC-support in $x_k$. As $x_j$ can be extracted from $L$ O$(d)$ times and each search from scratch costs O$(d)$, the total cost of checking for a PC-witness in $D(x_j)$, including the checks of line 4 in \textit{checkPCwit}, will be O$(d + d^2)$. For $d$ values of $x_i$ this will be O$(d^3)$. As this process will be repeated for all triangles of variables, whose number is bounded by $en$, its total cost will be O$(end^3)$. If no PC-witness is found then a new PC-support for $a_i$ in $D(x_k)$ is seeked through {\em searchPCwit}. This costs O$(nd^2)$ as explained above but it is amortized with the cost incurred by the calls to {\em searchPCwit} from {\em searchPCsup}. Therefore, the cost of {\em checkPCwit} is O$(end^3)$. This is also the node complexity of \texttt{maxRPC3}. 

The branch complexity of \texttt{(l)maxRPC3} is also O$(end^3)$. This is because the use of $LastPC$ ensures that for any constraint $c_{i,j}$ and a value $a_i\in D(x_i)$, each value of $x_j$ will be checked at most once for PC-support while going down the branch. Therefore, the cost of $searchPCwit$ is amortized. 
\end{proof}
\label{maxRPC3-complexity}
\end{proposition}

\begin{proposition}\rm
The node and branch time complexities of \texttt{lmaxRPC3$^{rm}$} and \linebreak \texttt{maxRPC3$^{rm}$} are O$(end^4)$ and O$(en^2d^4)$ respectively.
\newline 
\\
\begin{proof}The proof is similar to that of Proposition~\ref{maxRPC3-complexity}. The main difference with \texttt{lmaxRPC3} is that since $lastPC$ is not updated incrementally, each time we seek a PC-support for a value $a_i\in D(x_i)$ in $x_j$, $D(x_j)$ will be searched from scratch in the worst case. This incurs an extra O$(d)$ cost to {\em searchPCsup} and {\em searchPCwit}. Hence, the node complexity of \texttt{lmaxRPC3$^{rm}$} is O$(end^4)$. Also, the total cost of {\em searchPCwit} in one node cannot be amortized. This means that the cost of {\em searchPCwit} within {\em checkPCwit} is O$(nd^2)$. Hence, the node complexity of \texttt{maxRPC3$^{rm}$} is O$(en^2d^4)$. The branch complexities are the same because the calls to {\em searchPCwit} are amortized. 
\end{proof}
\end{proposition}

The space complexities of the algorithms are determined by the space required for data structures $LastPC$ and $LastAC$. Since both require O$(ed)$ space, this is the node space complexity of \texttt{(l)maxRPC3} and \texttt{(l)maxRPC3$^{rm}$}. \texttt{(l)maxRPC3} has O$(end)$ branch space complexity because of the extra space required for the incremental update and restoration of the data structures. As \texttt{(l)maxRPC3$^{rm}$} avoid this, its branch space complexity is O$(ed)$.

\section{Heuristics for maxRPC Algorithms}

\label{sec:heuristics}

Numerous heuristics for ordering constraint or variable revisions have been proposed and used within AC algorithms \cite{wallace92,boussem04,balsterg08}. %More generally, many constraint solvers employ heuristics to order the application of propagators or/and the revision of variables and constraints \cite{schulte08}. 
Heuristics such as the ones used by AC algorithms can be also used within a maxRPC algorithm to efficiently select the next variable to be removed from the propagation list (line 5 of Algorithm \ref{algMaxRPC3}). In addition to this, maxRPC and lmaxRPC algorithms can benefit from the use of heuristics elsewhere in their execution. Once a variable $x_j$ has been removed from the propagation list, heuristics can be applied as follows in either a maxRPC or a lmaxRPC algorithm (we use algorithm \texttt{(l)maxRPC3} for illustration):

\begin{enumerate}

\item After a variable $x_j$ is removed from $L$ all neighboring variables $x_i$ are revised. %A value $a_i\in D(x_i)$ is pruned if it has no PC-support in $D(x_j)$. If the value is not pruned then a maxRPC algorithm will check for PC-witness loss. % if the unique PC-support $a_k\in D(x_k)$ of $a_i$, where $x_k$ forms a triangle with $x_i$ and $x_j$, has a PC-witness in $D(x_j)$. 
lmaxRPC (resp. maxRPC) will detect a failure if the condition of PC-support loss (resp. either PC-support or PC-witness loss) occurs for all values of $x_i$. In such situations, the sooner $x_i$ is considered and the failure is detected, the more constraint checks will be saved. Hence, the order in which the neighboring variables of $x_j$ are considered can be determined using a fail-first type of heuristic. 

\item Once an AC-support $a_j\in D(x_j)$ has been found for a value $a_i\in D(x_i)$, {\em searchPCsup} tries to establish if it is a PC-support. % by searching for a PC-witness for the pair $(a_i,a_j)$ in the domains of all variables constrained with both $x_i$ and $x_j$. 
If there is no PC-witness for the pair $(a_i,a_j)$ in some variable $x_k$ then $a_j$ is not a PC-support. % and we can move on to check the next value in $D(x_j)$. 
%Hence, the sooner $x_k$ is considered, the more constraint checks will be saved. 
Therefore, we can again use fail-first heuristics to determine the order in which the variables forming a triangle with $x_i$ and $x_j$ are considered. 

\end{enumerate}  

The above cases apply to both lmaxRPC and maxRPC algorithms. In addition, a maxRPC algorithm can employ heuristics as follows:

\begin{enumerate}[start=3]

\item For each value $a_i \in D(x_i)$ and each variable $x_k$ constrained with both $x_i$ and $x_j$, Function \ref{algWitness} checks %if $a_k=LastPC_{x_i,a_i,x_k}$ is still valid and if it is, 
if the pair $(a_i,a_k)$ still has a PC-witness in $D(x_j)$. If there is no PC-witness or $LastPC_{x_i,a_i,x_k}$ is not valid then a new PC-support in $x_k$ is seeked. If none is found then $a_i$ will be deleted. %The sooner $x_k$ is considered, the more constraint checks will be saved which means that 
Again heuristics can be used to determine the order in which the variables constrained with $x_i$ and $x_j$ are considered. 

\item In Function \ref{algWitness} if $LastPC_{x_i,a_i,x_k}$ is not valid then a new PC-support for $a_i$ in $D(x_k)$ is seeked. The order in which variables constrained with both $x_i$ and $x_k$ are considered can be determined heuristically as in Case 2 above.

\end{enumerate}  

As explained, the purpose of such ordering heuristic will be to ``fail-first''. %\cite{haral80}. 
That is, to quickly discover potential failures (Case 1 above), refute values that are not PC-supports (Cases 2 and 4) and delete values that have no PC-support (Case 3). Such heuristics can be applied in any coarse-grained maxRPC algorithm to decide the order in which variables are considered in Cases 1-4. Examples are the following:

\begin{description}

\item[dom] Consider the variables in ascending domain size. This heuristic can be applied in any of the four cases. 

\item[del$\_$ratio] Consider the variables in ascending ratio of the number of remaining values to the initial domain size. This heuristic can be applied in any of the four cases. 

\item[wdeg] In Case 1 consider the variables $x_i$ in descending weight for the constraint $c_{ij}$. In Case 2 consider the variables $x_k$ in descending average weight for the constraints $c_{ik}$ and $c_{jk}$. Similarly for Cases 3 and 4. 

\item[dom/wdeg] Consider the variables in ascending value of dom/wdeg. This heuristic can be applied in any of the four cases.

\end{description}

%Specifically for algorithms \texttt{maxRPC2} and \texttt{maxRPC3}, we can also apply the following heuristics: 

%\begin{description}

%\item[dom$_>$LastPC] In Case 3 consider the variables $x_k$ in ascending number of remaining values that are lexicographically after $LastPC_{x_i,a_i,x_k}$. 

%\item[dom$_>$LastAC] In Case 2 consider the variables $x_k$ in ascending number of remaining values that are lexicographically after max($LastAC_{x_i,a_i,x_k},LastAC_{x_j,a_j,x_k})$. Similarly for Case 4. This heuristic is specific to algorithm \texttt{maxRPC3}.

%\end{description}

Experiments demonstrated that applying heuristics in Cases 1 and 3 are particularly effective, while doing so in Cases 2 and 4 saves constraint checks but only marginally reduces cpu times. All of the heuristics mentioned above for Cases 1 and 3 offer cpu gains, with dom/wdeg being the most efficient. Although the primal purpose of the heuristics is to save constraint checks, it is interesting to note that some of the heuristics can also divert search to different areas of the search space when a variable ordering heuristic like dom/wdeg is used, resulting in fewer node visits.
For example, two different orderings of the variables in Case 1 may result in different constraints causing a failure. As dom/wdeg increases the weight of a constraint each time it causes a failure and uses the weights to select the next variable, this may later result in different branching choices. This is explained for the case of AC in \cite{balsterg08}.

\section{Experiments}

\label{sec:experiments}

We have experimented with several classes of structured and random binary CSPs taken from C.Lecoutre's XCSP repository. Excluding instances that were very hard for all algorithms, our evaluation was done on 200 instances in total from various problem classes. More details about these instances can be found in C.Lecoutre's homepage. All algorithms used the dom/wdeg heuristic for variable ordering \cite{bhls04} and lexicographic value ordering. In case of a failure (domain wipe-out) the weight of constraint $c_{ij}$ is updated (right before returning in line 15 of Algorithm~\ref{algMaxRPC3}). The suffix '+H' after any algorithm's name means that we have applied the dom/wdeg heuristic for ordering the propagation list \cite{balsterg08}, and the same heuristic for \textit{Case 1} described in Section 4. In absense of the suffix, the propagation list was implemented as a FIFO queue and no heuristic from Section 4 was used.

\begin{table*}[hbt]
\centering\small
\begin{footnotesize}
\caption{Average stand-alone performance in all 200 instances grouped by problem class. Cpu times (t) in secs and constraint checks (cc) are given.} 

%\vspace{-3mm}
\begin{center}
\begin{scriptsize}
\begin{tabular}[hbt]{|@{~}l@{~}|@{~}c@{~}|@{~}c@{~}|@{~}c@{~}|@{~}c@{~}|@{~}c@{~}|@{~}c@{~}|@{~}c@{~}|@{~}c@{~}|}

\hline

Problem class & & maxRPC2 & maxRPC3 & lmaxRPC2 & lmaxRPC3 & lmaxRPC$^{rm}$ &  lmaxRPC3$^{rm}$ &  lmaxRPC3+H  \\
\hline

RLFAP & t & 6.786 & 2.329 & 4.838 & \textbf{2.043} & 4.615 & 2.058 & 2.148  \\
(scen,graph)       & cc & 31M & 9M & 21M & 8M & 21M & 9M & 8M \\

\hline

Random & t & 0.092 & 0.053 & 0.079 & 0.054 & 0.078 & \textbf{0.052} & 0.056  \\
(modelB,forced)       & cc & 0.43M & 0.18M & 0.43M	 & 0.18M  & 0.43M  & 0.18M & 0.18M \\

\hline

Geometric & t & 0.120 &	0.71 & 0.119 &	0.085 & 0.120 & 0.086 & \textbf{0.078} \\
       & cc & 0.74M & 0.35M & 0.74M	 & 0.35M  & 0.74M  & 0.35M & 0.35M  \\

\hline
Quasigroup & t & 0.293 & 0.188 & 0.234 & 0.166 & 0.224 & \textbf{0.161} & 0.184 \\
(qcp,qwh,bqwh)       & cc & 1.62M & 0.59M &	1.28M & 0.54M & 1.26M & 0.54M & 0.54M  \\

\hline

QueensKnights,   & t & 87.839 & 47.091 &	91.777 & 45.130 & 87.304 & 43.736 & \textbf{43.121}  \\
Queens,QueenAttack       & cc & 489M &	188M & 487M & 188M & 487M & 188M & 188M \\

\hline

driver,blackHole & t &	0.700 &	0.326 & 0.630 & \textbf{0.295} & 0.638 & 0.303 & 0.299 \\
haystacks,job-shop  & cc & 4.57M & 1.07M & 4.15M & 1.00M  & 4.15M & 1.00M & 1.00M  \\

\hline

\end{tabular}
\end{scriptsize}
\end{center}
%\vspace{-2mm}
\label{table:preprocessSummary}
\end{footnotesize}
\end{table*}

Table~\ref{table:preprocessSummary} compares the performance of stand-alone algorithms used for preprocessing. We give average results for all the instances, grouped into specific problem classes. We include results from the two optimal coarse-grained maxRPC algorithms, \texttt{maxRPC2} and  \texttt{maxRPC3}, from all the light versions of the coarse-grained algorithms, and from one of the most competitive algorithms (\texttt{maxRPC3}) in tandem with the dom/wdeg heuristics of Section \ref{sec:heuristics} (\texttt{lmaxRPC3+H}). Results show that in terms of run time our algorithms have similar performance and are superior to existing ones by a factor of two on average. This is due to the elimination of many redundant constraint checks as the cc numbers show. Heuristic do not seem to make any difference.

Tables ~\ref{table:searchRLFAP} and ~\ref{table:searchOther} compare the performance of search algorithms that apply \texttt{lmaxRPC} throughout search on RLFAPs and an indicative collection of other problems$\ $ respectively.$\ $ The$\ $ algorithms$\ $ compared $\ $are \texttt{\ lmaxRPC$^{rm}$\ } and \texttt{\ lmaxRPC3$^{rm}$} with and without the use of heuristic dom/wdeg for propagation list and for Case 1 of Section \ref{sec:heuristics}. We also include results from MAC$^{rm}$ which is considered the most efficient version of MAC \cite{lecoutre07,liki07}.

\begin{table*}[hbt]
\centering
\begin{footnotesize}
\caption{Cpu times (t) in secs, nodes (n) and constraint checks (cc) from RLFAP instances. Algorithms that use heuristics are denoted by their name + H. The best cpu time among the lmaxRPC methods is highlighted.} %\vspace{-3mm}
\begin{center}
\begin{scriptsize}
\begin{tabular}[hbt]{|@{~}l@{~}|@{~}c@{~}|@{~}c@{~}|@{~}c@{~}|@{~}c@{~}|@{~}c@{~}|@{~}c@{~}|}

\hline

instance & & AC$^{rm}$ & lmaxRPC$^{rm}$ & lmaxRPC3$^{rm}$ & lmaxRPC$^{rm}$ + H & lmaxRPC3$^{rm}$ + H \\
\hline

scen11 & t & 5.4 &	13.2 &	4.6 & 12.5 & {\bf 4.3} \\
       & n & 4,367 &	1,396 & 1,396 & 1,292 &	1,292 \\
       & cc & 5M & 92M & 29M & 90M & 26M \\

\hline

scen11-f10 & t & 11.0 & 29.0 &	12.3 &	22.3 &	{\bf 9.8} \\
       & n & 9,597 &	2,276 & 2,276 & 1,983 & 1,983 \\
       & cc & 11M & 141M & 51M & 114M & 41M \\

\hline

%scen11-f12 & t & 7,466 & 21,795 & 9,469 &	17,490 &	{\bf 8,144} \\
%       & n & 6,932 &	1,783 &	1,783 &	1,679	& 1,679 \\
%       & cc & 7.8M &	112M & 41M & 94M & 35M \\

%\hline

scen2-f25 & t & \textbf{27.1} & 109.2 & 43.0 & 79.6 & 32.6 \\
       & n & 43,536 & 8,310 & 8,310	& 6,179	& 6,179 \\
       & cc & 44M & 427M &	151M & 315M & 113M \\

\hline

scen3-f11 & t & \textbf{7.4} &	30.8 &	12.6 &	17.3 &	7.8 \\
       & n & 7,962 &	2,309 & 2,309 & 1,852 &	1,852 \\
       & cc & 9M & 132M & 46M & 80M & 29M \\

\hline

scen11-f7 & t & 4,606.5 & 8,307.5 & 3,062.8 & 6,269.0 & {\bf 2,377.6} \\
       & n & 3,696,154 & 552,907 & 552,907 &	522,061 & 522,061 \\
       & cc & 4,287M & 35,897M &	9,675M &	22,899M & 6,913M \\

\hline

scen11-f8 & t &\textbf{ 521.1} & 2,680.6 & 878.0 &	1,902.4 &	684.7 \\
       & n & 345,877 &	112,719 &	112,719 &	106,352 &	106,352 \\
       & cc & 638M &	10,163M &	3,172M &	7,585M &	2,314M \\

\hline

graph8-f10 & t & 16.4 & 16.8 & 9.1 &	11.0 &	{\bf 6.3} \\
       & n & 18,751 & 4,887 &	4,887 & 3,608 & 3,608 \\
       & cc & 14M & 71M & 31M & 51M & 21M \\

\hline

%graph8-f11 & t & 2,186 &	3,364 & 1,763 & 2,953 &	{\bf 1,685} \\
%       & n & 2,110 &	611 &	611 &	455 &	455 \\
%       & cc & 2.7M &	28M &	12M &	25M &	11M \\

%\hline

graph14-f28 & t & 31.4 & 4.1 & 3.1 &	2.6 & {\bf 2.1} \\
       & n & 57,039 & 2,917 &	2,917 & 1,187 & 1,187 \\
       & cc & 13M & 17M & 8M & 13M & 6M \\

\hline

graph9-f9 & t & 273.5 &	206.3 &	{\bf 101.5} &	289.5 &	146.9 \\
       & n & 273,766 &	26,276 &	26,276 &	49,627 &	49,627 \\
       & cc & 158M &	729M &	290M &	959M &	371M \\

\hline

%graph9-f10 & t & 12,906 &	20,320 &	9,781 &	13,644 &	{\bf 6,880} \\
%       & n & 9,396 &	1,351 &	1,351 &	1,149 &	1,149 \\
%       & cc & 11M &	102M &	42M &	73M &	29M \\

%\hline

%{\bf g7-w1-f4} & t & 252 & 135 & 125 & 212 & 196 \\
%       & n & 542 &	414 &	414 &	414 &	414  \\
%       & cc & 675K & 2.2M & 1.2M &	2.2M & 1.2M \\

%\hline

\end{tabular}
\end{scriptsize}
\end{center}
%\vspace{-2mm}
\label{table:searchRLFAP}
\end{footnotesize}
\end{table*}

Experiments showed that \texttt{lmaxRPC$^{rm}$} is the most efficient among existing algorithms when applied during search, which confirms the results given in \cite{vion09}. Accordingly, \texttt{lmaxRPC3$^{rm}$} is the most efficient among our algorithms. It is between two and four times faster than \texttt{maxRPC3$^{rm}$} on hard instances, while algorithms \texttt{lmaxRPC3} and \texttt{lmaxRPC2} are not competitive when used during search because of the data structures they maintain. In general, when applied during search, any maxRPC algorithm is clearly inferior to the corresponding light version. The reduction in visited nodes achieved by the former is relatively small and does not compensate for the higher run times of enforcing maxRPC.

Results from Tables \ref{table:searchRLFAP} and~\ref{table:searchOther} demonstrate that \texttt{lmaxRPC3$^{rm}$} always outperforms \texttt{lmaxRPC$^{rm}$}, often considerably. This was the case in all 200 instances tried. The use of heuristics improves the performance of both lmaxRPC algorithms in most cases. Looking at the columns for \texttt{lmaxRPC$^{rm}$} and \texttt{lmaxRPC3$^{rm}$} +H we can see that our methods can reduce the numbers of constraint checks by as much as one order of magnitude (e.g. in quasigroup problems qcp and qwh). This is mainly due to the elimination of redundant checks inside function {\em searchPCwit}. Cpu times are not cut down by as much, but a speed-up of more than 3 times can be obtained (e.g. scen2-f25 and scen11-f8). 

\begin{table*}[hbt]
\centering
\begin{footnotesize}
\caption{Cpu times (t) in secs, nodes (n) and constraint checks (cc) from various instances. %Instances where lmaxRPC is very close or better than AC are highlighted with bold. The best cpu time among the lmaxRPC methods is also highlighted.} 
}
%\vspace{-3mm}
\begin{center}
\begin{scriptsize}
\begin{tabular}[hbt]{|@{~}l@{~}|@{~}c@{~}|@{~}c@{~}|@{~}c@{~}|@{~}c@{~}|@{~}c@{~}|@{~}c@{~}|}

\hline

instance & & AC$^{rm}$ & lmaxRPC$^{rm}$ & lmaxRPC3$^{rm}$ & lmaxRPC$^{rm}$ + H & lmaxRPC3$^{rm}$ + H \\
\hline

rand-2-40-8 & t & \textbf{4.0} & 47.3 & 21.7 & 37.0 & 19.0 \\
-753-100-75  & n & 13,166 & 8,584 &	8,584 & 6,915 & 6,915 \\
       & cc & 7M & 289M & 82M & 207M &	59M \\

\hline

geo50-20 & t &\textbf{ 102.7 }& 347.7 & 177.5 &	273.3 & 150.3 \\
d4-75-1  & n & 181,560 & 79,691 & 79,691 & 75,339 & 75,339 \\
       & cc & 191M &	2,045M &	880M & 1,437M & 609M \\

\hline

%geo50-20 & t & 21,870 &	72,056 &	38,675 &	62,086 &	{\bf 36,114} \\
%d4-75-8  & n & 48,766 &	20,609 &	20,609 &	20,808 &	20,808 \\
%       & cc & 36M & 389M &	166M & 297M & 124M \\

%\hline

qcp150-120-5 & t & 52.1 & 89.4 & {\bf 50.2} & 80.0 & 55.3 \\
		 & n & 233,311 & 100,781 &	100,781 & 84,392 & 84,392  \\
       & cc & 27M & 329M &	53M &	224M & 36M \\

\hline

qcp150-120-9 & t & 226.8 & 410.7 &	238.1 & 239.9 & {\bf 164.3} \\
		 & n & 1,195,896 & 583,627 & 583,627 &	315,582 & 315,582 \\
       & cc & 123M &	1,613M &	250M & 718M & 112M \\

\hline

qwh20-166-1 & t & 52.6 &	64.3 &	38.9 &	21.2 &	{\bf 14.9} \\
		 & n & 144,653 & 44,934 &	44,934 &	13,696 &	13,696 \\
       & cc & 19M &	210M & 23M & 53M & 6M \\

\hline

%qwh20-166-5 & t & 28,159 &	26,402 &	15,787 &	23,646 &	{\bf 17,316} \\
%		 & n & 75,248 & 18,231 & 18,231 & 16,880 & 16,880 \\
%       & cc & 11M & 86M	& 10M & 58M & 7M \\

%\hline

qwh20-166-6 & t & 1,639.0 &	1,493.5 &	867.1 &	1,206.2 &	{\bf 816.5} \\
		 & n & 4,651,632 &	919,861 & 919,861 &	617,233 &	617,233 \\
       & cc & 633M &	5,089M &	566M & 3,100M & 351M \\

\hline

%qwh20-166-7 & t & 102,314 & 121,488 & {\bf 72,977} & 168,491 &	116,907 \\
%		 & n & 263,713 &	76,624 &	76,624 &	89,050 &	89,050 \\
%       & cc & 41M	& 392M &	45M &	430M & 48M \\

%\hline

%qwh20-166-8 & t & 299,474 & 457,786 & 270,223 &	361,190 & {\bf 273,905} \\
%		 & n & 864,657 &	271,602 &	271,602 &	211,241 & 211,241 \\
%       & cc & 117M &	1,509M &	169M & 851M &	96M \\

%\hline

qwh20-166-9 & t & 41.8 &	41.1 &	{\bf 25.0} &	39.9 &	28.5 \\
		 & n & 121,623 &	32,925 &	32,925 &	26,505 &	26,505 \\
       & cc & 15M &	135M &	15M &	97M &	11M \\

\hline

%bgwh15-106-4 & t & 164 & 106 & {\bf 83} &	291 &	340 \\
%		 & n & 2,088 &	477 &	477 &	1,354 &	1,354 \\
%       & cc & 126,720 &	362,067 & 161,652 & 746,923 &	292,774 \\

%\hline

blackHole & t &\textbf{ 1.8} &	14.4 &	3.8 &	12.1 &	3.6 \\
4-4-e-8	 & n & 8,661 &	4,371 &	4,371 &	4,325 &	4,325 \\
       & cc &  4M &	83M &	12M &	68M &	10M \\

\hline

queens-100 & t & \textbf{15.3} & 365.3 &	106.7 & 329.8 & 103.0 \\
	 & n & 7,608 &	6,210 &	6,210 &	5,030 &	5,030 \\
       & cc & 6M & 1,454M & 377M &	1,376M &	375M \\

\hline

queenAttacking5 & t & \textbf{34.3} & 153.1 &	56.7 &	136.0 & 54.8 \\
	 & n & 139,534 &	38,210 &	38,210 &	33,341 &	33,341 \\
       & cc & 35M &	500M &	145M &	436M &	128M \\

\hline

%queenAttacking6 & t & 207,932 &	1,808,610 &	570,024 & 1,435,741 & {\bf 482,118} \\
%	 & n & 262.087 &	103,058 & 103,058 &	98,504 &	98,504 \\
%       & cc & 210M & 6,034M &	1,640M &	4,726M &	1,297M \\

%\hline

queensKnights & t & 217.0 &	302.0 &	{\bf 173.6} &	482.0 &	283.5 \\
-15-5-mul	 & n & 35,445 &	13,462 &	13,462 &	12,560 &	12,560 \\
       & cc & 153M &	963M &	387M &	1,795M &	869M \\

\hline

%haystacks-05 & t & 1,500 &	993 &	714 &	2,165 &	1,942 \\
%	 & n & 110,638 &	20,278 &	20,278 &	62,478 &	62,478 \\
%       & cc & 1.4M &	1.9M & 1M &	4M &	2.5M \\

%\hline

\end{tabular}
\end{scriptsize}
\end{center}
%\vspace{-2mm}
\label{table:searchOther}
\end{footnotesize}
\end{table*}

Importantly, the speed-ups obtained can make a search algorithm that efficiently applies lmaxRPC competitive with MAC on many instances. For instance, in scen11-f10 we achieve the same run time as MAC while \texttt{lmaxRPC$^{rm}$} is 3 times slower while in scen11-f7 we go from 2 times slower to 2 times faster. In addition, there are several instances where MAC is outperformed (e.g. the graph RLFAPs and most quasigroup problems). Of course, there are still instances where MAC remains considerably faster despite the improvements. 

\begin{table*}[htb]
\centering
\begin{footnotesize}
\caption{Average search performance in all 200 instances grouped by class.} 

%\vspace{-2mm}
\begin{center}
\begin{scriptsize}
\begin{tabular}[hbt]{|@{~}l@{~}|@{~}c@{~}|@{~}c@{~}|@{~}c@{~}|@{~}c@{~}|@{~}c@{~}|@{~}c@{~}|}

\hline

Problem class & & AC$^{rm}$ & lmaxRPC$^{rm}$ & lmaxRPC3$^{rm}$ & lmaxRPC$^{rm}$ + H & lmaxRPC3$^{rm}$ + H \\
\hline

RLFAP & t & 242.8 & 556.7 & 199.3 & 416.3 & \textbf{157.3} \\
%(scen,      & n & 195,160 & 33,152 & 33,152 & 32,074 & 32,074 \\
(scen,graph)       & cc & 233M & 2,306M & 663M & 1,580M &	487M \\

\hline

Random & t & \textbf{8.4} & 28.0 & 14.8 &	28.5 & 17.1 \\
%(modelB,       & n & 31,615 & 14,489 & 14,489 & 14,622 & 14,622 \\
(modelB,forced)       & cc & 14M &	161M &	60M & 137M & 51M \\

\hline

Geometric & t & \textbf{21.5} &	72.2 &	37.2 &	57.6 &	32.1 \\
%       & n & 39,879 &	17,273 &	17,273 &	16,567 &	16,567 \\
       & cc & 39M & 418M &	179M & 297M & 126M \\

\hline
Quasigroup & t & 147.0 & 162.5 & 94.9 & 128.9 & \textbf{89.6} \\
%(qcp,		 & n & 456,324 & 123,478 &	123,478 & 83,615 & 83,615  \\
(qcp,qwh,bqwh)       & cc & 59M & 562M &	68M &	333M & 40M \\

\hline

QueensKnights,   & t & \textbf{90.2} & 505.2 &	180.3 & 496.4 & 198.1 \\
%Queens,		 & n & 67,015 & 24,854 & 24,854 &	23,169 & 23,169 \\
Queens,QueenAttack       & cc & 74M &	1,865M &	570M & 1,891M & 654M \\

\hline

driver,blackHole & t &	\textbf{3.2} &	17.1 & 9.1 & 11.9 & 7.0 \\
%(driver,blackHole,	& n & 17,427 & 6,911 &	6,911 &	9,428 &	9,428 \\
haystacks,job-shop  & cc & 1.8M &	55M & 6.4M & 36.7M & 5.1M \\

\hline

\end{tabular}
\end{scriptsize}
\end{center}
%\vspace{-2mm}
\label{table:searchSummary}
\end{footnotesize}
\end{table*}

Table~\ref{table:searchSummary} summarizes results from the application of lmaxRPC during search. We give average results for all the tested instances, grouped into specific problem classes. As can be seen, our best method improves on the existing best one considerably, making lmaxRPC outperform MAC on the RFLAP and quasigroup problem classes. Overall, our results demonstrate that the efficient application of a maxRPC approximation throughout search can give an algorithm that is quite competitive with MAC on many binary CSPs. This confirms the conjecture of \cite{db01} about the potential of maxRPC as an alternative to AC. In addition, our results, along with ones in \cite{vion09}, show that approximating strong and complex local consistencies can be very beneficial.

\section{Conclusion}

\label{sec:conclusion}

We presented \texttt{maxRPC3} and \texttt{maxRPC3$^{rm}$}, two new algorithms for maxRPC, and their light versions that approximate maxRPC. These algorithms build on and improve existing maxRPC algorithms, achieving the elimination of many redundant constraint checks. We also investigated heuristics that can be used to order certain operations within maxRPC algorithms. Experimental results from various problem classes demonstrate that our best method, \texttt{lmaxRPC3$^{rm}$}, constantly outperforms existing algorithms, often by large margins. Significantly, the speed-ups obtained allow \texttt{lmaxRPC3$^{rm}$} to compete with and outperform MAC on many problems. In the future we plan to adapt techniques for using residues from \cite{liki07} to improve the performance of our algorithms during search. Also, it would be interesting to investigate the applicability of similar methods to efficiently achieve or approximate other local consistencies.

\bibliographystyle{plain}
\begin{footnotesize}
\bibliography{maxRPC}
\end{footnotesize}

\end{document}